\documentclass[11pt]{article}
\usepackage{acl2015}
\usepackage{times}
\usepackage[T1,hyphens]{url}
\usepackage{latexsym}
\usepackage{makecell}
\usepackage{tabularx}
\usepackage[english]{babel}
\usepackage{biblatex}
\usepackage{booktabs}
\usepackage{hyperref}
\usepackage{xurl}

\title{JUSTICE: A Benchmark Dataset for Supreme Court's Judgment Prediction}

\author{Mohammed Alsayed, Shaayan Syed, Mohammad Alali, Smit Patel, Hemanth Bodala \\
  University of Southern California \\
  {\tt \{alsayedm, shaayans, alalim, smitp, bodala\}@usc.edu}}

\begin{document}
\maketitle

\section{Introduction}

Over 7,000 Supreme Court cases are requested to be reviewed each year in the United States; however, only 100 cases on average are heard. \cite{scotus_yearly} Most of this data is raw, and while there are existing datasets comprised of this kind of information, they were of other countries' court cases such as the Chinese Supreme Court and the European Human Rights court. \cite{xiao2018cail2018, Aletras2016-tu} There was no existing dataset that was created to perform the task we have outlined in this paper on the Supreme Court of the United States.

Our goal is to compile a well-annotated dataset that can be readily used in natural language processing (NLP) and other data-driven applications. While court cases have an abundance of information detailing them, the ones that concern us the most include the absolute neutral facts of the case and the details of the decisions made by the presiding justices, such as the voting counts and the winning party.

Furthermore, we train a model to predict a court's judgment based on the facts provided by both the petitioner and respondent. In other words, the model is simulating a human jury by generating a final verdict. Recently, advancements in NLP and machine learning provide us with the tools to build predictive models that can be used to reveal patterns that influence Supreme Court case decisions. This modeling problem has the potential to be solved using the dataset we have created.

Datasets similar to this already exist. However, as pointed out earlier, they are directed towards a specific category of cases, such as those involving antitrust laws, rather than the entire spectrum we wish to analyze. One related research paper that is an example of this is titled “CAIL2018: A Large-Scale Legal Dataset for Judgement Prediction” by \citeauthor{xiao2018cail2018}. \cite{xiao2018cail2018} Their goal of predicting judgement results is comparable to ours, but the big difference is that their 2.6 million cases dataset only consists of criminal cases. \cite{xiao2018cail2018} In addition, these cases were published by the Supreme People’s Court of China which are vastly greater in number than the cases in our dataset which pertain to the Supreme Court of the United States.

Given that there have been a significant number of Supreme Court cases filed over the years, we looked to the databases hosted by the Caselaw Access Project and Oyez as our source of raw information. Due to the challenges involving missing data and a scarcity of relevant entries, we also had to locate other sources to patch these holes; the methodology for this will be detailed further on. Following that, we trained a few models to gauge the dataset's versatility and quality. By analyzing the results of this study, we gained a better understanding of how the dataset could be modified to improve and train more accurate models in the future.

\begin{table*}[ht]
\begin{tabularx}{\textwidth}{l|X}
\hline
\bf Column & \bf Description \\ \hline
ID & Unique case identifier. \\ \hline
Name & The name of the case. \\ \hline
HREF & The Oyez’s API URL for the case. \\ \hline
Docket ID & A special identifier of the case used by the legal system. \\ \hline
Term & The year when the Court received the case. \\ \hline
First Party & The name of the first party (petitioner). \\ \hline
Second Party & The name of the second party (respondent). \\ \hline
Facts & The absolute, neutral facts of the case written by the court clerk. \\ \hline
Majority Vote & The number of justices voting for the majority opinion. \\ \hline
Minority Vote & The number of justices voting for the minority opinion. \\ \hline
Winning Party & The name of the party that won the case. \\ \hline
First Party Winner & True if the first party won the case, otherwise False and the second party won the case. \\ \hline
Decision Type & The type of the decision decided by the court, e.g.: per curiam, equally divided, opinion of the court. \\ \hline
Disposition & The treatment the Supreme Court accorded the court whose decision it reviewed; e.g.: affirmed, reversed, vacated. \\ \hline
Issue Area & The pre-defined legal issue category of the case; e.g.: Civil Rights, Criminal Procedure, Federal Taxation. \\ \hline
\end{tabularx}
\caption{\label{table-dataset-def} Dataset columns definition. }
\end{table*}

\section{Methods}

\subsection{Dataset Preparation}

A large portion of our time was focused on researching which source of Supreme Court cases would provide the most appropriate data for our purposes. Given our goal, we needed to find one that provided a large enough text summary of the facts of each case, as well as the final decision made by the judges. Additional information regarding the case decision was welcome. Our top two candidates were the Caselaw Access Project (CAP) and Oyez. While both sources contain the facts of the case and the outcome, the facts from CAP were usually short and contained little-to-no information to base a decision on. Hence, due to the fact that Oyez had richer facts and a simple API to retrieve cases, we decided to use it as our main source of data.

Unfortunately, a major downside of Oyez was that only a total of 8,209 cases were available, which was reasonable considering that only 100 cases are heard annually by the Supreme Court. \cite{scotus_yearly} Nevertheless, we ran into further issues with this database which reduced this number even further.

\subsubsection{Dataset Definition}

The most important part of a dataset is the way it's structured and the data it contains. Hence, after much deliberation, we landed into the dataset with the following columns as listed in Table~\ref{table-dataset-def} after filtering the data from Oyez. Additionally, we kept the Oyez URL that the case was obtained from for future reference.

\subsubsection{Data Complications}

As mentioned earlier, the biggest problem we encountered was the fact that a significant portion of the cases we collected had either missing or incorrect information. The following are examples of the issues found while analyzing the dataset. Firstly, there were some ambiguous winning party names that could not be identified whether it’s a substring of the first or second party as shown in Table~\ref{table-ambigious}.

\begin{table}[htb]
\centering
\setlength\tabcolsep{0pt}
\begin{tabular*}{\linewidth}{@{\extracolsep{\fill}} *{4}{c}}
\toprule
\textbf{Case Name} & \textbf{Party1} & \textbf{Party2} & \textbf{Winner} \\
\midrule
Boggs v. Boggs & Boggs & Boggs & Boggs \\
\bottomrule
\end{tabular*}
\caption{\label{table-ambigious} Case with ambiguous winning party. }
\end{table}

Secondly, the dataset had swapped values between the majority and minority votes for some instances since it didn’t make sense to have a larger minority vote than a majority vote. An example of a swapped vote is illustrated below in Table~\ref{table-swapped-vote}.

\begin{table}[ht]
\centering
\setlength\tabcolsep{0pt}
\begin{tabular*}{\linewidth}{@{\extracolsep{\fill}} *{3}{c}}
\toprule
\textbf{Case Name} & \textbf{Majority Vote} & \textbf{Minority Vote} \\
\midrule
Taylor v. Barkes & 0 & 9 \\
\bottomrule
\end{tabular*}
\caption{\label{table-swapped-vote} Case with incorrectly swapped votes. }
\end{table}

Thirdly, some of the data had an unmatching decision type with its voting values. An example of a case where its votes do not reflect the decision type is shown in Table~\ref{table-wrong-decision-type}.

\begin{table}[ht]
\centering
\setlength\tabcolsep{0pt}
\begin{tabular*}{\linewidth}{@{\extracolsep{\fill}} *{4}{c}}
\toprule
\textbf{Name} & \textbf{Maj. Vote} & \textbf{Min. Vote} & \textbf{Decision} \\
\midrule
Sears vs. LA & 0 & 8 & equally divided \\
\bottomrule
\end{tabular*}
\caption{\label{table-wrong-decision-type} Case with incorrect decision type. }
\end{table}

Finally, some of the cases had a total of zero votes. All of these issues have been solved by comparing them with the Washington University Law database \url{http://scdb.wustl.edu/} which is a more robust database of the supreme court that contains solely the decisions. However, some cases were not found so they have been kept as is or dropped if they lacked information, and the ones matched have been updated with the correct values.

\subsection{Dataset Preprocessing}

\subsubsection{Data Cleaning}

Once the dataset has been populated, the next step was to enrich the data by removing unnecessary information to ensure a high-quality dataset. Primarily, the data cleaning process was focused on the textual aspect of the dataset, which included the party names and the facts. First, all HTML tags and URLs were removed from the text as it served no significance to the models utilizing the dataset. Second, the excess spaces and all non-UTF8 characters were removed. Third, the contractions in the dataset were expanded to their normal form; for example, the word \textit{“can’t”} was converted into \textit{“cannot”}. Finally, the stopwords in text, such as \textit{“the”}, were discarded as they added hardly any information. This process not only improved the quality of text in the dataset by discarding irrelevant text, it also shrank the size of the average facts by 20\% which could facilitate faster learning and processing.

\subsubsection{Data Augmentation}

Even though we managed to preserve most of our dataset, we still had a limited number of cases to work with. Furthermore, we observed that in a majority of our entries, the judges favored the second party in only one out of eight cases, while the rest of our entries showed a majority of the judges favoring the first party; i.e. a 88-12 class imbalance in the dataset. To combat this problem, we have augmented the cases where the second party won by generating similar cases. Data augmentation was performed using the \texttt{nlpaug} library, which provides textual augmentation for machine learning applications. There are three levels of data augmentation, i.e. characters, words, and sentences. In this work, we performed data augmentation on words, by replacing some of the words in the text with their synonyms or similar words.

The first approach was back-translation, where a phrase is translated into another language and translated back to English. This approach replaced very few words in the sentence, and we found that the augmented sentences were nearly the same as the original ones. In the second technique, we replaced the words with their synonyms, but it still did not allow us to create a new sentence that maintained the meaning but was different syntactily. In the last two techniques, we used word embeddings to find the words which are similar to the word and replaced it; out of these two techniques, we adapted the contextual word embeddings technique which was the best one among the ones we tried. This method not only replaces the word based on similarity but also substitutes the best-fit words based on the contextual word embeddings; using this technique, we successfully replaced 50\% of the words in the facts without affecting the nouns. Using the bert-base-cased pre-trained model, we replaced the words with their four most similar words from the embeddings. This model has the advantage of persisting the case sensitivity of our data and the essence of the case while paraphrasing the facts into something novel. This process was repeated until the dataset had a 60-40 class balance.

\subsubsection{Data Mirroring}

While the Data Augmentation step lessened the dataset imbalance issue, we complemented it with another procedure to mitigate the imbalanced data even further. We noticed that the two classes in the dataset were interchangeable; class 1 is when the first party wins and class 2 is when the second party wins. Consequently, being the first party or the second party has no bearing on the outcome of the case. Thus, we mirrored each case in the dataset, where the first and second party names were swapped. Not only did this double our dataset’s size and end the class imbalance by having a 50-50 class balance, it also emphasized that the outcome of a case is invariant to the position of the parties.

\section{Experiments}

While our main priority was for the dataset to be used to predict the final decision for a Supreme Court case, we prepared several models and data to also solve another task as well in order to demonstrate its versatility. So along with the decision prediction task, we also used the dataset to train a model that predicts how controversial each court case would be.

\subsection{Decision Prediction Task}

The primary task to evaluate the dataset was to predict the judicial judgment given the absolute neutral facts of the case written by the court clerks. The judgment was broken into a binary classification; the winner of the case was mapped to 0 when it was the first party, or 1 otherwise. To accomplish this task, instead of having a single model predict the outcome, we opted for an ensemble model which encompasses multiple diverse models that each vote for an outcome and then the majority vote wins; exactly the same way the Supreme Court works with 9 justices. However, due to time limitations, we trained only 7 models to be a part of the ensemble model rather than the thematic 9. The models used are perceptron, support vector machines (SVM), logistic regression, naive bayes, multi-layer perceptron (MLP), k-nearest neighbors (KNN), and calibrated classifier. The \texttt{scikit-learn} package was utilized to train the models, complemented by the usage of GridSearchCV to obtain decent hyperparameters, which are listed in Table~\ref{table-hyperparameters}.

\begin{table}[ht]
\centering
\setlength\tabcolsep{0pt}
\begin{tabular*}{\linewidth}{@{\extracolsep{\fill}} *{2}{l}}
\toprule
\textbf{Model} & \textbf{Hyperparameters} \\
\midrule
SVM & \texttt{max\_iter=5} \\
\hline
Logistic Regression & \texttt{default} \\
\hline
Naive Bayes & \texttt{alpha=3} \\
\hline
Calibrated Classifier & \texttt{default} \\
\hline
Perceptron & \makecell[l]{\texttt{max\_iter=5} \\ \texttt{penalty='l1'}} \\
\hline
KNN & \makecell[l]{\texttt{n\_neighbors=3} \\ \texttt{weights='distance'}} \\
\hline
MLP & \makecell[l]{\texttt{early\_stopping=True} \\ \texttt{beta\_2=0} \\ \texttt{max\_iter=10}} \\
\bottomrule
\end{tabular*}
\caption{\label{table-hyperparameters} Hyperparameters for the models used in the paper. }
\end{table}

As the models do not explicitly take in raw textual format, we had to preprocess all textual input and vectorize it. We used the TF-IDF vectorizer to transform the document into an array of numbers which is more suitable for the model’s input. TF-IDF is a statistical measurement used to weigh words by their importance to a document; however, it does not take into account the context nor its position in the document. \cite{tfidf} Next, the models were fed the vectorized first party name, second party name, and the facts as input. The party names were a part of the input so that the model can distinguish between the entities mentioned in the facts of the case; furthermore, since there is a mirrored case where the party names were swapped around, the model should learn that the party name’s position has no bearing on the outcome. Moreover, other models utilizing LSTMs were attempted in our experiments; however, due to the size of the facts for each case, which on average is 189 words after preprocessing, it became difficult to correctly predict and train due to the vanishing and/or exploding gradients. \cite{lstm}

\subsection{Controversialness Prediction Task}

Another task we liked to accomplish with the constructed dataset was to determine, based on the facts presented, if a particular case would be controversial among the judges. In this context, we determine how controversial a case is by how close the final decisions by the Supreme Court Justices are to an even split. We determine whether a case is controversial based on the following expression: $1-\frac{majority\ votes}{total\ votes}$, where the class 1 represents cases with a controversy score of 0.3 or greater. The threshold value 0.3 was chosen to get a roughly even split of controversial to non-controversial cases.

Once every case is assigned a score representing how controversial it is, the next step is to generate word embeddings for the entire corpus of facts in each case. We construct a Word2Vec model based on the facts of all cases we have available, and do so with minimal preprocessing to ensure a model that is as large as possible. We opted to construct a completely novel Word2Vec model rather than use a premade model using data from search engines or other text sources because we believed that having a model constructed out of legal text would capture lexical similarity better in this context specifically. The model was constructed with a vector size of 100 items and a window size of 11. Once the Word2Vec embeddings were constructed, we have both an input and labels and can now begin training models to classify how controversial a case may be.

For this instance, we constructed both an SVM classifier as well as an LSTM to determine the performance of the dataset using different classifiers. To train the SVM, we calculated the average Word2Vec vector for the facts of every case and used that as input for the classifier, with the labels calculated earlier as our targets. To train the LSTM, we passed an array of vectors for each word of the facts of a case. Every array of vectors was padded to match the length of the longest set of facts that existed in our dataset.

\section{Results}

\subsection{Decision Prediction Task}

Before data augmentation and case mirroring, we had 2384 cases of which are 2114 class 1 (first party won) and 270 class 2 (second party won). Data augmentation was done only on class 2 which increased the dataset size to 3464 cases of which are 2114 class 1 and 1350 class 2. Next, each case was mirrored which doubled the number of cases to 6928 while having 3464 of each case; a 50-50 perfectly balanced dataset. This case mirroring ensured us that the position of the party is invariant to the outcome. Hence, while these data augmentation methods synthetically alleviated the class imbalance issue, it is still preferable to obtain a dataset that is naturally balanced; however, that is rarely the case. Table~\ref{table-results} below depicts the result of the models against a testing dataset, which was a 20\% split of the entire dataset. The models were evaluated using three metrics: accuracy (Acc\%), macro precision (MP\%), macro recall (MR\%), and macro F1 score (MF1). Collectively, these metrics should give us an insight of the performance of the models.

\begin{table}[htb]
\centering
\setlength\tabcolsep{0pt}
\begin{tabular*}{\linewidth}{@{\extracolsep{\fill}} *{5}{c}}
\toprule
\textbf{Model} & \textbf{Acc\%} & \textbf{MP\%} & \textbf{MR\%} & \textbf{MF1} \\
\midrule
Perceptron & 65\% & 65\% & 65\% & 0.65 \\
SVM & 60\% & 60\% & 60\% & 0.60 \\
Logistic Reg & 61\% & 61\% & 61\% & 0.61 \\
Naive Bayes & 59\% & 59\% & 59\% & 0.59 \\
MLP & 64\% & 64\% & 64\% & 0.64 \\
KNN & 68\% & 69\% & 68\% & 0.67 \\
Calib. Classifier & 62\% & 63\% & 62\% & 0.62 \\
Ensemble & 62\% & 62\% & 62\% & 0.62 \\
\bottomrule
\end{tabular*}
\caption{\label{table-results} Evaluation metric for each trained model. }
\end{table}

As shown in Table ~\ref{table-results}, the ensemble achieved an average accuracy of 62\%. It is decent enough to show that the model is not randomly choosing between the classifications with a 50\% chance for each. However, we were unable to achieve an accuracy higher than this throughout the variations of models that we experimented with. This shows us that the problem most likely lies in the dataset. The data augmentation methods we used earlier improved the problems of the limited entries, as well as the imbalance towards one type of classification, but it did not completely solve the dilemma. We hypothesize that this could be accounted for by incorporating other sources of Supreme Court cases in order to fill out this dataset with a more balanced ratio, thus giving us a higher accuracy if used to train these models again.

The lack of data and balanced outcomes played a part in the inability to provide compelling results for the Controversial task. Although, an additional problem also lies in the content of the input itself. The summarized facts of the court cases occasionally provided a comprehensive look at the case, but a significant amount of them contained almost no information at all. The average character length was 1,197, with the minimum being 95 and the maximum being 6108. For word length, the average measurement was 189 words. Given this knowledge, and the fact that the minimum number of words was 13, we can see that the summarized facts have very little content.

\subsection{Controversialness Prediction Task}

Unfortunately, we were unable to produce meaningful results from either the LSTM or the SVM in completing the task to determine if a case is controversial or not. The SVM classifier was able to produce an accuracy of ~61\%, however, upon further inspection, this is because the model will guess that every case is non-controversial. And since the non-controversial cases make up approximately 60\% of the dataset, this accuracy value is meaningless. As for the LSTM, we were unable to train the model in a way that produced quality results. After training the model with all of the facts for several hundred epochs with various hyper-parameter configurations, we were still unable to improve the model’s cross entropy loss past 0.6.

We believe that these results came about due to a flaw in the task itself. In other text-based binary labelling tasks such as sentiment analysis, there are clear targets for a model to pick up in order to make an accurate prediction. In the case of sentiment analysis these can be words such as ‘outstanding’ or ‘terrible’. These features are all present within the input text to begin so there is no external knowledge needed. However, for legal cases, the facts of the case itself may not immediately give hints as to how controversial a case may be. Typically, a case is controversial due to prior cases and laws that set a precedence favoring one party over another. This means to get an accurate picture of how controversial a case is, we must train it with extensive knowledge about prior legal cases and laws that could impact the case at hand. This is no fault of the dataset in itself, but rather a fault of a task that could never be accomplished with this dataset alone.

\section{Conclusions}

Based on the results produced by the models described above, we have concluded that there are still improvements to be made to this dataset. First and foremost is to increase the number of entries that are included in the data. This can be done by searching through other sources for the Supreme Court cases in order to account for incorrect or incomplete data. Another important priority is to make sure the dataset is not heavily skewed towards one specific outcome. Even though the data imbalance issue is lessened by the data augmentation and data mirroring procedures mentioned above, it is still preferable to have novel data that is naturally balanced. Again, this can be accounted for by increasing the number of entries.

The versatility of this database should also be considered when updating it, which is why one improvement could also be to split the facts of the case based on whether it corresponds to the first or second party. This type of format could be useful for a number of varying models, one example being to predict which party was the petitioner and which one was the respondent. Similar adjustments to the dataset like this one would no doubt broaden its usage.

An assumption we made during the data preparation stage was that the size of the summarized facts would be sufficient for our models to accurately predict the outcomes. Unfortunately, it is difficult to increase the length without using artificial augmentation methods. One solution we can use to improve the content itself, however, is to manually annotate the text, similar to named-entity recognition (NER) tagging. For example, we would replace every reference to the first party in the text with \texttt{<p1>}, and the second party with \texttt{<p2>}. This has the potential to work especially well considering that we primarily used BERT to obtain the word embeddings, which bases the embedding on the context in which the word is used. Furthermore, removing the party names from the learning ensures that the model is learning to deduce the correct judgement based solely on the facts of the case and less about who is in the case.

Given all these adjustments, the next step would be to test this updated dataset on the same models used in this report so that a clear improvement, or decline, can be ascertained.

\hypersetup{ breaklinks=true, colorlinks=true, pdfusetitle=true }
\printbibliography

@misc{scotus_yearly,
    title={Granted \& noted list: October term 2020 cases for argument},
    author={Supreme Court of the United States},
    year={2020}
}

@ARTICLE{Aletras2016-tu,
  title     = "Predicting judicial decisions of the European Court of Human
               Rights: a Natural Language Processing perspective",
  author    = "Aletras, Nikolaos and Tsarapatsanis, Dimitrios and Preo{\c
               t}iuc-Pietro, Daniel and Lampos, Vasileios",
  journal   = "PeerJ Comput. Sci.",
  publisher = "PeerJ",
  volume    =  2,
  number    = "e93",
  pages     = "e93",
  month     =  oct,
  year      =  2016,
  language  = "en"
}

@misc{xiao2018cail2018,
      title={CAIL2018: A Large-Scale Legal Dataset for Judgment Prediction}, 
      author={Chaojun Xiao and Haoxi Zhong and Zhipeng Guo and Cunchao Tu and Zhiyuan Liu and Maosong Sun and Yansong Feng and Xianpei Han and Zhen Hu and Heng Wang and Jianfeng Xu},
      year={2018},
      eprint={1807.02478},
      archivePrefix={arXiv},
      primaryClass={cs.CL}
}

@article{tfidf,
author = {Ramos, Juan},
year = {2003},
month = {01},
pages = {},
title = {Using TF-IDF to determine word relevance in document queries}
}

@article{lstm,
   title={Fundamentals of Recurrent Neural Network (RNN) and Long Short-Term Memory (LSTM) network},
   volume={404},
   ISSN={0167-2789},
   url={http://dx.doi.org/10.1016/j.physd.2019.132306},
   DOI={10.1016/j.physd.2019.132306},
   journal={Physica D: Nonlinear Phenomena},
   publisher={Elsevier BV},
   author={Sherstinsky, Alex},
   year={2020},
   month={Mar},
   pages={132306}
}

\end{document}